\documentclass[runningheads]{llncs}
\usepackage[T1]{fontenc}
\usepackage{graphicx}
\usepackage{caption,subcaption}
\usepackage{multirow}
\usepackage[table]{xcolor}
\usepackage{url}
\usepackage{color}
\begin{document}
\title{Explaining Recovery Trajectories of Older Adults Post Lower-Limb Fracture Using Modality-wise Multiview Clustering and Large Language Models}
%
%
\author{Shehroz S. Khan\inst{1,2}\orcidID{0000-0002-1195-4999} \and
Ali Abedi\inst{1,3}\orcidID{0000-0002-7393-1362} \and
Charlene H. Chu\inst{3}\orcidID{0000-0002-0333-7210}}
\authorrunning{Khan et al.}
%
\institute{KITE Research Institute, Toronto Rehabilitation Institute, University Health Network, Toronto, Canada 
\and
College of Engineering and Technology, American University of the Middle East, Egaila, Kuwait
\and
Lawrence Bloomberg Faculty of Nursing, University of Toronto, Toronto, Canada\\
\email{\{shehroz.khan, ali.abedi\}@uhn.ca} \email{charlene.chu@utoronto.ca}
}
\maketitle              
\begin{abstract}
Interpreting large volumes of high-dimensional, unlabeled data in a manner that is comprehensible to humans remains a significant challenge across various domains. In unsupervised healthcare data analysis, interpreting clustered data can offer meaningful insights into patients' health outcomes, which hold direct implications for healthcare providers. This paper addresses the problem of interpreting clustered sensor data collected from older adult patients recovering from lower-limb fractures in the community. A total of 560 days of multimodal sensor data, including acceleration, step count, ambient motion, GPS location, heart rate, and sleep, alongside clinical scores, were remotely collected from patients at home. Clustering was first carried out separately for each data modality to assess the impact of feature sets extracted from each modality on patients' recovery trajectories. Then, using context-aware prompting, a large language model was employed to infer meaningful cluster labels for the clusters derived from each modality. The quality of these clusters and their corresponding labels was validated through rigorous statistical testing and visualization against clinical scores collected alongside the multimodal sensor data. The results demonstrated the statistical significance of most modality-specific cluster labels generated by the large language model with respect to clinical scores, confirming the efficacy of the proposed method for interpreting sensor data in an unsupervised manner. This unsupervised data analysis approach, relying solely on sensor data, enables clinicians to identify at-risk patients and take timely measures to improve health outcomes.

\end{abstract}
\vspace*{0.01in}
\section{Introduction}
Clustering is the process of grouping data objects such that the objects within the same clusters are more similar to each other than to the objects in other clusters \cite{khan2004cluster}. The majority of clustering algorithms use some notion of similarity to group different objects into distinct clusters. Based on the data type (numerical, categorical, or mixed), similarity or dissimilarity can be calculated using many available metrics, such as Euclidean distance, Hamming distance, Jaccard coefficient, and other specialized metrics \cite{xu2005survey}. Several types of clustering algorithms exist, including partitional, hierarchical, density-based, spectral, subspace, and others \cite{xu2005survey,xu2015comprehensive}.
In multiview clustering approaches \cite{chao2021survey}, the data are clustered based on 'views' or subspaces of the data, where a view contains a specific aspect of information (e.g., data collected from different sensors makes different views of information). Then, the data are clustered using those views and eventually merged to give final clusters using various approaches \cite{yang2018multi}.

Irrespective of the clustering approach, the overall clustering process is objective, data-driven, and unsupervised, i.e., it groups data objects using some notion of (dis)similarity, mostly without any context, which results in cluster labels that are largely arbitrary. Human involvement is needed to understand the contents of these clusters to determine what each cluster represents. In a healthcare setting, these clusters could represent healthy, non-healthy, or at-risk patients. This would be vital information for artificial intelligence (AI)-driven healthcare decision support systems, enabling clinicians to intervene promptly and improve patient outcomes.


It is well known that in high-dimensional multimodal data, different data modalities or features extracted from different data modalities interact with each other in complex ways, making it difficult to explain the clusters obtained through a clustering algorithm. Therefore, interpreting clusters is a challenging problem. Recent developments in Large Language Models (LLMs), e.g., ChatGPT \cite{mann2020language}, DeepSeek \cite{bi2024deepseek}, provide an opportunity to analyze and query multimodal datasets following clustering. These models can assist in generating human-interpretable explanations for cluster structures that are otherwise difficult to interpret \cite{hasasneh2023wearable,gao2024unsupervised,10.1145/3696461,verma2025enhanced,paraschou2024human,ke2024integrating}.
This paper presents a novel multiview clustering pipeline in which data are clustered based on specific data modalities, and cluster labels are inferred through context-aware LLM prompting. To evaluate the proposed method, a multimodal sensor dataset (MAISON-LLF) \cite{abedi2025maison} was used, comprising 560 days' worth of data, including acceleration, step count, ambient motion, GPS location, heart rate, and sleep, collected from ten community-dwelling older adult patients recovering from lower-limb fractures. The data were continuously collected using a smartwatch, smartphone, and sleep and motion sensors. Using the proposed multiview clustering–LLM pipeline, the contribution of each modality to the recovery trajectory was contextualized. The MAISON-LLF dataset \cite{abedi2025maison} is also accompanied by a range of clinical scores, such as social isolation and other metrics related to functional recovery. Through rigorous statistical analysis and visualization, it was shown that most of the derived cluster labels are in agreement with clinical scores, thereby reaffirming the validity of the proposed approach. The low statistical significance of a few clusters underscores the importance of improved feature extraction methods to more effectively capture and explain patient health outcomes, suggesting that domain-informed feature engineering or representation learning techniques may further enhance the interpretability and clinical relevance of clustering results.

The paper is structured as follows: Section \ref{sec:literature_review} reviews related LLM-based cluster interpretation methods. Section \ref{sec:method} details the proposed methodology. Section \ref{sec:experiments} presents experimental settings and results. Section \ref{sec:conclusion} concludes and outlines future work.

\vspace*{0.01in}
\section{Literature Review}
\label{sec:literature_review}
This section reviews related work on unsupervised sensor data analysis through clustering and the use of LLMs. The primary rationale behind these approaches is to enable interpretation of sensor data without manual effort. LLMs can serve as a valuable tool for labeling clustered sensor data. Typically, the sensor data is first clustered algorithmically, and then an LLM is prompted to assign labels or meaningful descriptions to the clusters or their centers. These LLM-generated labels are subsequently propagated to the data samples within each cluster for further analysis.

Hasasneh et al. \cite{hasasneh2023wearable} proposed an unsupervised learning framework using smartwatch sensor data, heart rate, heart rate variability, and step count, for COVID-19 detection and monitoring. By applying clustering techniques like K-means and DBSCAN, the framework identifies health states based on deviations from baseline physiology. An explainability layer using Decision Trees and GPT-3 Davinci \cite{brown2020language} provides interpretable summaries. Evaluated on 28 participants (11 COVID-positive), the approach effectively detects infection-related changes, correlating with symptom onset and severity. Results show that unsupervised clustering outperforms human-labeled data, reducing the need for supervised annotations in early infection detection.

Gao et al. \cite{gao2024unsupervised} proposed LLMs and Iterative Evolution for Unsupervised Human Activity Recognition (LLMIE-UHAR), an unsupervised human activity recognition framework that combines LLMs with iterative evolution. The method begins by applying K-means clustering to extract key representative sensor samples, which are then converted into textual prompts enriched with contextual and semantic information. LLMs use these prompts to annotate activities without manual labeling. A CNN model is trained on the annotated samples and refined iteratively. Evaluated on the Activity Recognition with Ambient Sensing dataset \cite{alemdar2013aras} using ambient sensor data, the method achieved 96\% accuracy across nine selected human activities, demonstrating its effectiveness in unsupervised activity recognition using structured LLM-based interpretation.

Hota et al. \cite{10.1145/3696461} investigated the feasibility of using LLMs as virtual annotators for time-series physical sensing data, specifically for human activity recognition, to eliminate the need for traditional human-in-the-loop annotation. They introduced a self-supervised learning-based encoding approach, where raw inertial measurement unit data is transformed into feature-rich embeddings, so-called clusters, using contrastive learning techniques before being fed into the LLM for annotation. These embeddings enhance the LLM's ability to classify human activities without fine-tuning or complex prompt engineering. The method was evaluated on four benchmark human activity recognition datasets. Results show that LLM-based annotation significantly improves human activity recognition accuracy when combined with contrastive learning-based embeddings, demonstrating that while LLMs alone are insufficient for direct sensor data analysis, integrating contrastive learning bridges this gap, making LLMs effective virtual annotators.

Natarajan et al. \cite{verma2025enhanced}, along with Khare and Kumar \cite{khare2023summarizing}, introduced a novel approach to interpreting clustered data in different domains using LLMs. The method derives key attributes from clusters by analyzing statistical properties such as cluster size, variance, density, spatial positioning, and inter-cluster relationships. These extracted keywords serve as structured inputs for the cluster interpreter, which transforms them into coherent linguistic descriptions of the clusters. The model is trained using T5 \cite{raffel2020exploring} and GPT \cite{brown2020language} architectures, both of which are fine-tuned to produce fluent, non-monotonous, and contextually rich cluster interpretations and summaries across various domains. The approach is evaluated using multiple real-world datasets, including the Iris dataset, airline dataset, and credit card dataset, where clustering algorithms such as K-means, DBSCAN, and single-linkage hierarchical clustering are applied. The study's results highlight that LLMs enhance the interpretability of clustering outcomes, making data-driven insights more accessible to domain non-experts.

Paraschou \cite{paraschou2024human} introduced a method that combines clustering, explainable AI, and LLMs to improve the interpretability of a complex dataset. The framework begins with clustering algorithms such as K-means, DBSCAN, and hierarchical clustering to group similar data points. Explainability techniques such as Shapley Additive Explanations (SHAP) \cite{lundberg2017unified} are then applied to identify influential features within each cluster. Finally, LLMs such as GPT-4 \cite{openai2023gpt4} and Mistral 7B \cite{jiang2023mistral} generate human-readable summaries that translate numerical patterns into intuitive insights. The approach was evaluated on the LifeSnaps dataset \cite{yfantidou2022lifesnaps}, a multimodal time-series dataset containing data modalities such as physical activity, sleep, heart rate variability, stress, and personality traits. Results demonstrate that LLM-based summaries significantly enhance the interpretation of sensor data related to physical and mental health, offering more accessible and meaningful analysis compared to traditional methods in healthcare and well-being monitoring.

Ke et al. \cite{ke2024integrating} proposed Integrate Vision-Language Semantic Graphs in multiview Clustering (IVSGMV), a novel method that integrates vision-language semantic graphs with multiview clustering to enhance performance on high-dimensional image datasets. Using Contrastive Language-Image Pretraining (CLIP) \cite{radford2021learning}, the method builds bipartite graphs linking image embeddings to text concepts from WordNet \cite{fellbaum1998wordnet}. An adaptive hybrid graph filter then balances homophily- and heterophily-based information to refine clustering. Evaluated on large-scale multiview datasets, IVSGMV outperforms traditional baselines and effectively leverages zero-shot capabilities of pre-trained models, achieving improved clustering without the need for labeled data.

Despite existing evidence in the literature highlighting the applicability and effectiveness of LLMs for labeling and interpreting clustered sensor data \cite{verma2025enhanced,khare2023summarizing,gao2024unsupervised,10.1145/3696461,ke2024integrating,hasasneh2023wearable,paraschou2024human}, particularly in the context of unsupervised remote patient monitoring \cite{hasasneh2023wearable,paraschou2024human}, none of the previous studies have specifically focused on understanding social isolation and functional decline among older adult patients discharged home after lower-limb fracture surgery. Moreover, no prior research has combined multimodal sensor data with corresponding multiview clustering and LLM-driven interpretation to provide a holistic view of behavioral patterns in this population. Given the unique and often varied patterns of daily activities, mobility, and lifestyle commonly observed among older adults, a clear research gap remains, especially regarding the integration of contextual information about this specific problem into LLM frameworks. To address this gap, this paper presents a novel approach that leverages multimodal sensor data with multiview clustering to identify patterns associated with the daily behaviors of older adults recovering from lower-limb fractures. This methodology is further enhanced by the use of problem-specific context-aware prompts, enabling LLMs to automatically label and explain the resulting clusters, supporting an unsupervised and comprehensive understanding of social isolation and functional decline. This approach minimizes dependence on subjective clinical data or patient self-reports, offering objective, scalable insights into the health and behavioral patterns of older adults. The quality of the inferred cluster labels was further validated using visualizations and statistical analysis of the corresponding clinical scales.

\vspace*{0.01in}
\section{Multiview Clustering and LLM}
\label{sec:method}
\begin{figure}[!ht]
    \centering
    \includegraphics[width=0.7\linewidth]{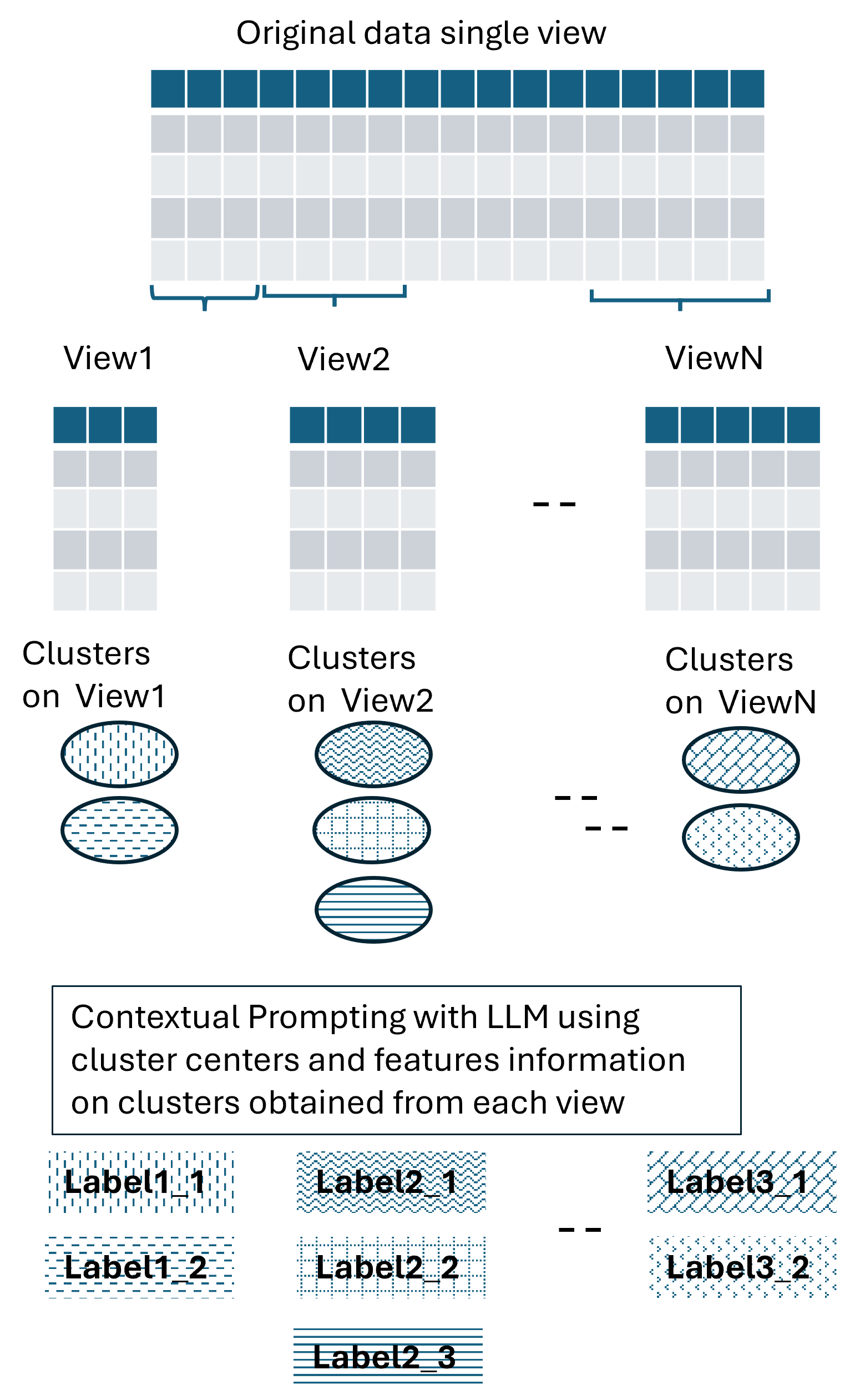}
    \caption{Generation of cluster labels on multimodal data using multiview clustering and context-aware prompts provided to LLMs.}
    \label{fig:pipeline}
\end{figure}

This section presents the idea of explainable multiview clustering aimed at interpreting multimodal data in an unsupervised manner. The dataset under consideration is called MAISON-LLF \cite{abedi2025multimodal}, which contains 560 days of sensor data across multimodal features collected from various sensors deployed in the homes of patients recovering from lower-limb fractures (see more details in Section \ref{sec:maison-llf}). A single-view clustering approach could result in multiple clusters across many features that may be difficult to interpret. To address this, the proposed method clusters the data by each modality and then uses context-aware prompting to generate explainable labels describing each modality-specific cluster. This approach facilitates understanding the role of each modality in mapping the recovery trajectory of patients once they return to the community after being discharged following a lower-limb fracture. Finally, statistical analysis is conducted to determine whether the clusters formed through the proposed explainable multiview clustering represent statistically significant clinical information. The overall pipeline is shown in Figure \ref{fig:pipeline}.

\vspace*{0.01in}
\subsection{MAISON-LLF dataset}
\label{sec:maison-llf}
The Multimodal AI-based Sensor platform for Older iNdividuals with Lower-Limb Fractures (MAISON-LLF) dataset is a publicly available, multimodal sensor dataset collected from older adults recovering from lower-limb fractures in community settings \cite{abedi2025multimodal}. It includes data from smartphone and smartwatch sensors, motion detectors, sleep-tracking mattresses, and clinical questionnaires related to social isolation and functional decline. The dataset was collected from 10 older adults living alone for eight weeks each, totaling 560 days of continuous, 24-hour monitoring. It spans six sensor modalities: acceleration, heart rate, step, GPS location, motion, and sleep, with a range of features extracted from each. Additionally, participants completed biweekly clinical assessments, including the Social Isolation Scale (SIS) \cite{nicholson2020psychometric}, Oxford Hip and Knee Scores (OHS, OKS) \cite{wylde2005oxford,whitehouse2005oxford}, the Timed Up and Go (TUG) test \cite{podsiadlo1991timed}, and the 30-Second Chair Stand Test \cite{chen2009normative}, providing ground-truth measures of social and functional health. The data collection process began shortly after discharge from the hospital. The dataset is publicly accessible via Zenodo \cite{abedi2025maison}, offering raw sensor data, extracted features, and detailed documentation in multiple formats.

\subsection{Data Processing}
\label{sec:dataprocessing}
The version of the MAISON-LLF dataset used in this paper is available in tabular format \cite{abedi2025multimodal}. It contains 560 rows, each representing a day of data, and 80 columns comprising daily clinical information and features extracted from multimodal sensors. Clinical assessments were conducted bi-weekly; however, all 14 preceding days were assigned the same values for clinical assessments. Moreover, the clinical information should not be confused as a feature, but rather considered ground truth in this context to validate the clustering of days. For simplicity, we refer to clinical information as features. Therefore, out of the 80 features, three, 'participant', 'timestamp', and 'clinical-timestamp', were deemed unnecessary and removed from the analysis. Of the remaining 77 features, 42 were clinical or demographic features, while 35 were numerical features representing data extracted from six data modalities on a daily basis. For clustering purposes, these modalities are referred to as 'views', as they represent different types of features extracted from patients. The position view included 3 features, the motion sensor view included 5, the heart rate view included 4 features, the sleep view included 11 features, the step view included 5 features, and the acceleration view included 7 features. The clinical and demographic features were excluded from further analysis, and only numerical features were used. This was done to investigate the role of sensor data in understanding the recovery trajectory of patients. The clinical features were used to assess the quality of the clusters in accurately grouping patients.

\vspace*{0.01in}
\subsection{Multiview Clustering and Cluster Labeling}
\label{sec:multiview_clustering}
K-means clustering was used to perform multiview clustering. The random state was set to a fixed value to ensure reproducibility of results. If the random state is changed to another value, the final clustering results (and cluster centers) could differ. All the data were normalized before clustering to prevent features with larger numerical ranges from disproportionately influencing the results. After clustering was completed, the cluster centers were computed and converted back to the original scale using reverse normalization. This facilitated the interpretation of features and the design of context-aware prompts for LLMs to infer relevant cluster labels.

\section{Experiments and Results}
\label{sec:experiments}

\begin{table}[!ht]
    \centering
    \caption{Two cluster centers obtained from the position view}
    \begin{tabular}{|c||c|c|c|} \hline
     \multirow{2}{4em}{\textbf{Cluster}} & \multicolumn{3}{|c|}{\textbf{Features}} \\ \cline{2-4}
     & \textbf{position-count} & \textbf{position-duration} & \textbf{position-travelled-distance} \\ \hline
    1 & 41.218925 & 94.892652 & 3.445000 \\
    2 & 16.500000 & 21.000000 & 3632.845000 \\ \hline
    \end{tabular}
    \label{tab:cc}
\end{table}

\subsection{Multiview Clustering and Cluster Labeling}
\label{sec:cluster_labeling}
Each view described in Section \ref{sec:dataprocessing} was clustered using the K-means clustering algorithm, varying the number of clusters from $K=2$ to $15$, and the best clusters were recorded for each data modality. The position, motion, step, and acceleration views each yielded 2 optimal clusters, whereas the heart rate view yielded 5, and the sleep view yielded 4 optimal clusters. As an example, the cluster centers generated by the position view are outlined in Table \ref{tab:cc}.

It can be observed that Cluster 1 represents days when patients traveled short distances but remained outside, whereas Cluster 2 represents days involving longer-distance travel. The following context-aware prompt was provided to the GPT-4o LLM to infer labels for each cluster, along with short descriptions:
\\
{\fontfamily{qcr}\selectfont

\noindent
I have clustered data into two clusters. Each cluster has three features, namely,
\begin{itemize}
    \item position-count - The total count of position data in a day.
    \item position-duration - The duration (in hours) of being outside the home in a day.
    \item position-travelled-distance - The total distance (in kilometers) traveled outside the home in a day.
\end{itemize}

The two cluster centers are as follows:
\begin{itemize}
    \item Cluster 1: 41.218925 94.892652 3.445000
    \item Cluster 2: 16.500000 21.000000 3632.845000
\end{itemize}
This data was collected from smartphones with GPS carried by different persons and features were extracted from it. These people are patients recovering from lower-limb fractures and were living at home. Please suggest cluster names for each cluster for better interpretation. Only provide a short cluster name and interpretation. Do not provide any additional analysis.
\\
}

The LLM provided the following suggestions:
\\
{\fontfamily{qcr}\selectfont
\begin{itemize}
    \item Cluster 1: \textit{Low Mobility with Extended Outdoor Time} – Patients spend a long duration outside but travel minimal distances, possibly engaging in sedentary or localized activities.
    \item Cluster 2: \textit{High Mobility with Long-Distance Travel} – Patients travel significant distances, indicating higher mobility and potentially advanced recovery or external assistance for transportation.
\end{itemize}

}

Using a similar two-step process, cluster labels were generated for each sensor data modality as shown in Table \ref{tab:clusterlabels}.

\begin{table}[!htbp]
    \centering
    \caption{Cluster labels and their short description obtained for each cluster (for each sensor view or modality) after performing K-means clustering and context-aware prompting with the GPT-4o LLM.}
    \resizebox{\textwidth}{!}{%
    \begin{tabular}{|p{50pt}|p{120pt}|p{220pt}|} \hline
    \textbf{View} & \textbf{Cluster Labels} & \textbf{Cluster Label Description} \\ \hline
      Motion & Active Home Movers   & Higher motion count and spread-out activity suggest frequent movement within the living space \\ 
         & Sedentary Home Dwellers & Lower motion count and reduced activity indicate a more stationary lifestyle with minimal movement \\ \hline
    Heart Rate & Moderate Activity with Stable Heart Rate & Balanced heart rate values with low variability \\
     & Low Activity with Anomalous Readings & Extremely low mean heart rate and missing minimum values \\
     & Elevated Heart Rate with Variability & Higher heart rate with noticeable fluctuations \\
     & Erratic Heart Rate Patterns & Very low mean heart rate with extreme variability \\
     & Resting State with Mild Fluctuations & Lower heart rate with minimal variation \\ \hline
    Sleep & Extended Sleep with Frequent Snoring & Long total sleep duration with notable snoring \\
     & Short Sleep with Low Disturbance & Reduced total sleep with minimal wake-ups \\
     & Light Sleep Dominant with High Wake-Ups & Predominantly light sleep with frequent wake-ups \\
     & Elevated Heart Rate with Moderate Sleep & Moderate sleep duration with high heart rate \\ \hline
     Step & Sedentary Routine & Lower step count and less frequent movement throughout the day \\
      & Frequent Movers & Higher step count and more consistent movement patterns \\ \hline
    Acceleration & Low Activity Movers & Lower acceleration count and variability, indicating limited movement \\
     & High Activity Movers & Higher acceleration count and variability, suggesting more frequent or intense movement \\ \hline
    \end{tabular}
    }
    \label{tab:clusterlabels}
\end{table}

\subsection{Clinical Validation}
\begin{figure}[!h]
    \centering 
\begin{subfigure}{0.5\textwidth}
  \includegraphics[width=\linewidth]{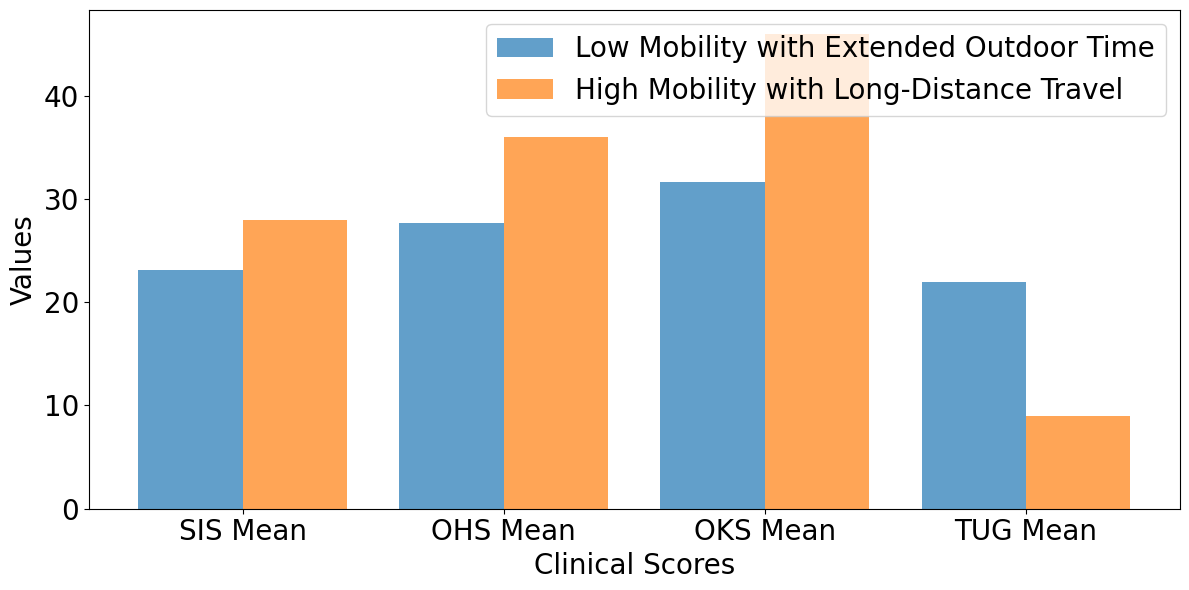}
  \caption{Position View}
  \label{fig:position}
\end{subfigure}\hfil 
\begin{subfigure}{0.5\textwidth}
  \includegraphics[width=\linewidth]{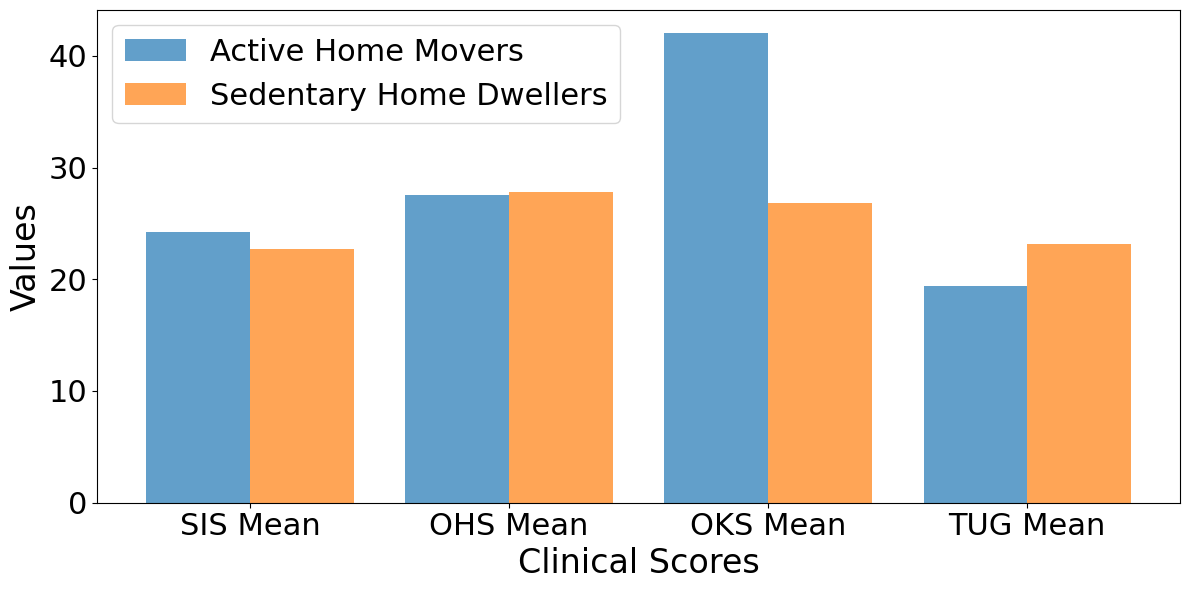}
  \caption{Motion View}
  \label{fig:motion}
\end{subfigure}\hfil 

\medskip
\begin{subfigure}{0.5\textwidth}
  \includegraphics[width=\linewidth]{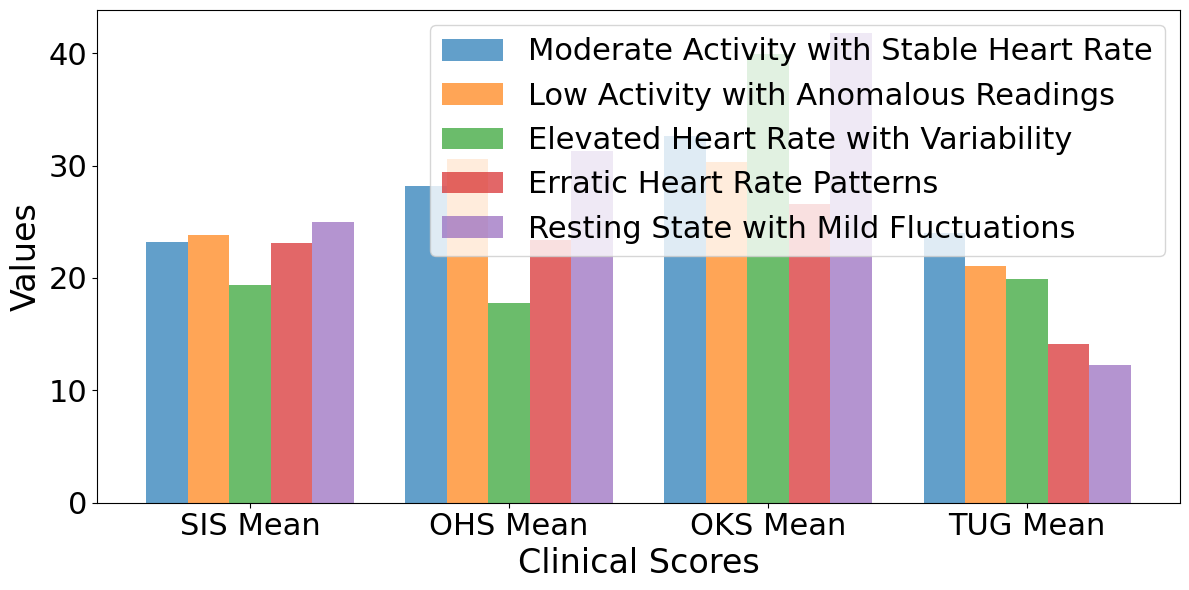}
  \caption{Heartrate View}
  \label{fig:heart}
\end{subfigure}\hfil
\begin{subfigure}{0.5\textwidth}
  \includegraphics[width=\linewidth]{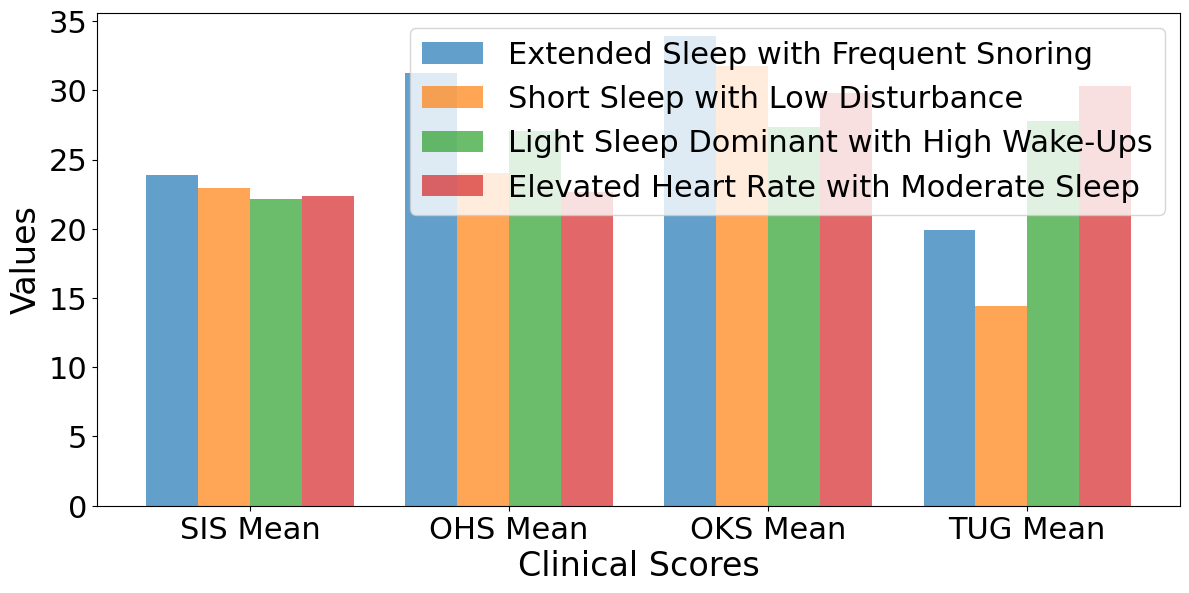}
  \caption{Sleep View}
  \label{fig:sleep}
\end{subfigure}\hfil 

\medskip
\begin{subfigure}{0.5\textwidth}
  \includegraphics[width=\linewidth]{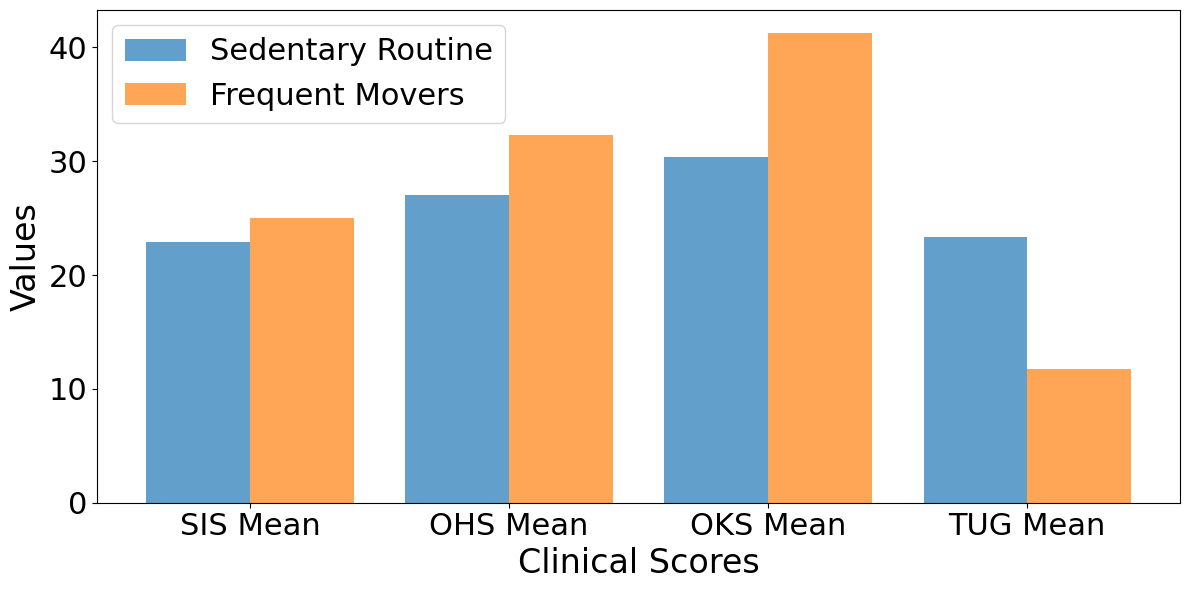}
  \caption{Step View}
  \label{fig:step}
\end{subfigure}\hfil 
\begin{subfigure}{0.5\textwidth}
  \includegraphics[width=\linewidth]{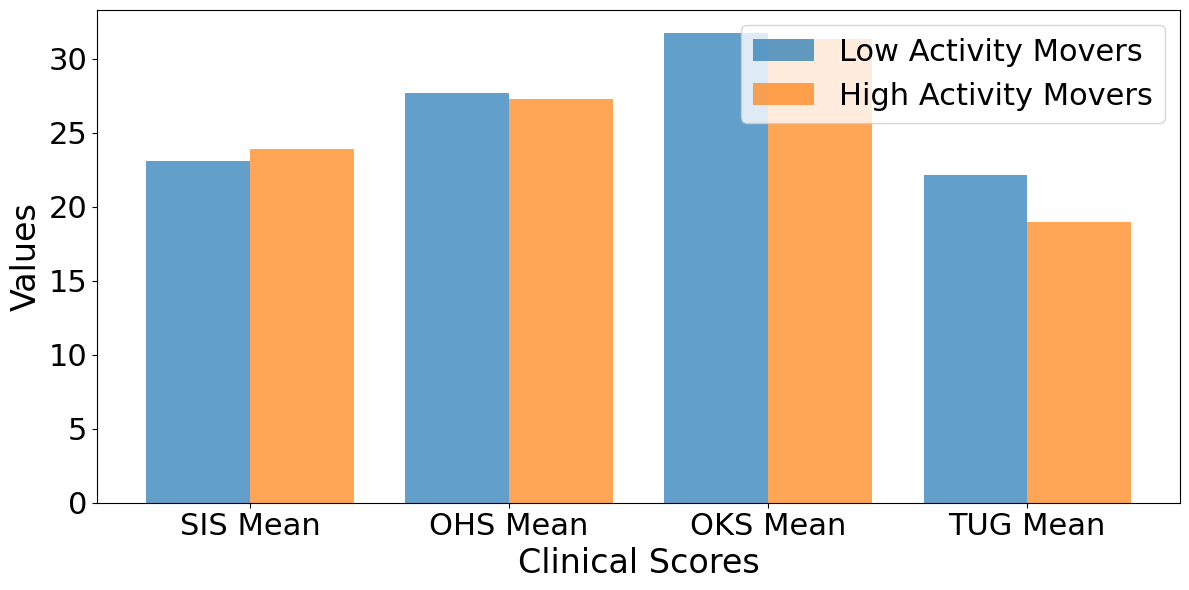}
  \caption{Acceleration View}
  \label{fig:acceleration}
\end{subfigure}
\caption{The mean distribution of various clinical scales for each clusters obtained from different views from the multimodal sensor data.}
\label{fig:clinicalscores}
\end{figure}

As described in subsection \ref{sec:maison-llf}, the MAISON-LLF dataset also contains bi-weekly clinical assessments for each patient. These assessments were applied to the dataset on a daily basis, such that a value collected on a given day was assigned to the preceding two weeks. The distribution of mean clinical scores for each cluster obtained by each sensor data modality is shown in Figure \ref{fig:clinicalscores}. As an example, the mean SIS, OHS, OKS, and TUG values for the two clusters from the position view are $23.16, 27.67, 31.65, 21.98$ and $28.0, 36.0, 46.0, 9.0$, respectively (see Figure \ref{fig:position}). Visually, these values differ for each clinical score. Referring back to the cluster labels obtained in the previous section, Clusters $1$ and $2$ were labeled as \textit{Low Mobility with Extended Outdoor Time} and \textit{High Mobility with Long-Distance Travel}, respectively. Thus, it can be anticipated that Cluster $1$ could exhibit higher social isolation, lower hip function, lower knee function, and reduced overall body function compared to Cluster $2$'s mean clinical values. This confirms that the cluster labels obtained are potentially meaningful representations of the information contained in each cluster from the position view. However, this distinction may not be as visually evident in other cases, such as the heart rate view (see Figure \ref{fig:heart}). The heart rate cluster centers had abnormal values due to occasional zero heart rate readings, which are not realistic and were caused by hardware limitations of the smartwatch.
Therefore, statistical tests were performed to determine whether the clinical scores corresponding to each cluster within a modality are significantly different from one another, as described in the following subsection.

\subsection{Statistical Testing}
Statistical tests were conducted to assess the significance of the distributions of SIS, OHS, OKS, and TUG clinical scores associated with the clusters obtained for each sensor data modality. If there were two clusters and both followed a normal distribution, a t-test was performed; otherwise, the Mann–Whitney U test (MWUT) was used. If there were more than two clusters and all followed a normal distribution, a repeated-measure ANOVA test was conducted; otherwise, the Kruskal–Wallis test (KWT) was used. Although the dataset included a limited number of participants, the number of samples per cluster was sufficient in most cases to meet the assumptions of the statistical tests. The null hypothesis was that there is no significant difference in clinical scores between clusters, i.e., the mean clinical scores of the clusters are the same. If the p-value was less than $0.05$, the null hypothesis was rejected, indicating a statistically significant difference in clinical scores between clusters.

Table \ref{tab:stat_testing} shows the statistical testing results for each cluster corresponding to each sensor data modality. 
The columns Cluster1-5 shows the number of clusters per modality. A dash (-) means the cluster is absent, for e.g., Position view has only two clusters, so there is a dash for Cluster3-5. In each of the cluster1-5 column, a Yes, No or NaN means whether the samples in a cluster are normally distributed or not or insufficient data (less than three samples) following the Shapiro-Wilk test. 
For clusters obtained from heart rate, sleep, and step data, there are statistically significant differences between clusters. This indicates that the sensor data was able to group samples into clusters with clinical values that were significantly different from one another. For the position clusters, OHS and OKS scores were not statistically significant, and for the motion clusters, OHS was not statistically significant. In the position view, the clusters were heavily skewed, with Cluster 1 and Cluster 2 containing 558 and 2 data points, respectively. As a result, the test could not compute reliable statistics (hence the NaN value in Table \ref{tab:stat_testing} corresponding to Cluster 1). Overall, statistical testing shows that the clusters contained samples that were statistically distinguishable based on the distribution of clinical scores. This further supports the validity of the cluster labels inferred in subsection \ref{sec:cluster_labeling}.

\begin{table}[!h]
    \centering
    \caption{Statistical testing results for each cluster corresponding to each sensor data modality. P-values less than 0.0001 are reported as "< 0.0001". All other p-values are rounded to four decimal places. SIS: Social Isolation Scale, OHS: Oxford Hip Score, OKS: Oxford Knee Score, TUG: Timed Up and Go test, MWUT: Mann–Whitney U Test, KWT: Kruskal–Wallis Test.}
    \resizebox{\textwidth}{!}{%
    \begin{tabular}{|p{50pt}|p{35pt}|c|c|c|c|c|c|c|p{50pt}|} \hline
    \textbf{View} & \textbf{Clinical Score} &  \textbf{Cluster1} & \textbf{Cluster2} & \textbf{Cluster3} & \textbf{Cluster4} & \textbf{Cluster5} & \textbf{Test Used}  &  \textbf{p-value} & \textbf{Statistical Significance} \\ \hline

 \multirow{2}{4em}{Position} & SIS & No &  NaN  & - & - & - & MWUT  & 0.0360   & \cellcolor{gray!25}Yes \\ \cline{2-10}
 & OHS & No &  NaN & - & - & - &  MWUT  & 0.1581   & No \\ \cline{2-10}
 & OKS &  No &  NaN & - & - & - &   MWUT  &  0.0703   & No \\ \cline{2-10}
 & TUG &  No & NaN & - & - & - & MWUT  & 0.0499   & \cellcolor{gray!25}Yes \\ \hline \hline

 \multirow{2}{4em}{Motion} & SIS & No &  No & - & - & - & MWUT & < 0.0001  & \cellcolor{gray!25}Yes  \\ \cline{2-10}
 & OHS & No & No & - & - & - & MWUT & 0.8137   & No \\ \cline{2-10}
 & OKS & No & No & - & - & - & MWUT & < 0.0001   & \cellcolor{gray!25}Yes \\ \cline{2-10}
 & TUG & No & No & - & - & - & MWUT  & 0.0006  & \cellcolor{gray!25}Yes \\ \hline \hline

 \multirow{2}{4em}{Heart Rate} & SIS & No & No & No & No & No &  KWT & < 0.0001 & \cellcolor{gray!25}Yes   \\ \cline{2-10}
 & OHS & No & No & No & No & No &  KWT & < 0.0001  & \cellcolor{gray!25}Yes   \\ \cline{2-10}
 & OKS & No & No & No & No & No  & KWT & < 0.0001  & \cellcolor{gray!25}Yes   \\ \cline{2-10}
 & TUG & No & No & No & No & Yes & KWT & 0.0013  & \cellcolor{gray!25}Yes   \\ \hline \hline

 \multirow{2}{4em}{Sleep} & SIS   & No & No & No & No & - &  KWT & < 0.0001   &  \cellcolor{gray!25}Yes \\ \cline{2-10}
 & OHS & No & No & No & No & - & KWT & < 0.0001    &  \cellcolor{gray!25}Yes \\ \cline{2-10}
 & OKS & No & No & No & No & - & KWT & < 0.0001  &  \cellcolor{gray!25}Yes \\ \cline{2-10} 
 & TUG & No & No & No & No  & - & KWT & < 0.0001 &  \cellcolor{gray!25}Yes  \\ \hline \hline

 \multirow{2}{4em}{Step} & SIS & No & No & - & - & - & MWUT & < 0.0001 & \cellcolor{gray!25}Yes \\ \cline{2-10}  
 & OHS & No & No & - & - & - & MWUT & < 0.0001 & \cellcolor{gray!25}Yes \\ \cline{2-10}  
 & OKS & No & No & - & - & - & MWUT & < 0.0001 & \cellcolor{gray!25}Yes \\ \cline{2-10}  
 & TUG & No & No & - & - & - & MWUT & < 0.0001 & \cellcolor{gray!25}Yes \\ \hline \hline

 \multirow{2}{4em}{Acceleration} & SIS & No & No & - & - & - & MWUT &  0.1020  & No \\ \cline{2-10}   
 & OHS & No & No & - & - & -  & MWUT &  0.8241  & No \\ \cline{2-10}     
 & OKS & No & No & - & - & -  & MWUT &  0.6818  & No \\ \cline{2-10}     
 & TUG & No & No & - & - & -  & MWUT &  0.3892  & No \\ \hline   
    \end{tabular}
    }
    \label{tab:stat_testing}
\end{table}

The clusters obtained from the acceleration view were not found to be statistically significant with respect to any of the clinical scores. This was also visually observed in Figure \ref{fig:acceleration}, where most of the cluster center values appear similar for each clinical score. Although movement is known to be directly correlated with improved functional recovery and reduced social isolation \cite{rosso2013mobility} (as shown by the results from the motion sensor view, see Figure \ref{fig:motion}), this result appears counterintuitive. It is likely due to limitations in the sensitivity and resolution of the wearable device in capturing nuanced movement and inconsistencies in participant usage with the wearable device.


A python notebook containing all the experiments presented in this paper is available at \url{https://github.com/shehrozskhan/InterpretMultiview}

\subsection{Limitations}
The results presented in this paper provide unique insights into the recovery trajectories of older adult patients following lower-limb fractures. However, the study has certain limitations:

\begin{itemize}
    \item All clustering results were obtained using a fixed random seed. The optimal number of clusters and their interpretations may vary slightly with different random initializations, as this could lead to a different clustering outcome.
    \item Cluster centers average the features of all data points, so the impact of distant points or outliers within a cluster may not be well represented.    
    \item LLMs can produce different cluster labels even when given the same prompts and cluster centers. However, it is expected that, despite variations in wording, the underlying meaning of the labels should remain consistent.
    \item The clusters obtained from the acceleration view were not found to be statistically significant, likely due to limitations in the study design that affected the consistency and richness of the data collected. The protocol may not have sufficiently accounted for variations in daily routines or provided enough structure to ensure uniform data capture across participants. As a result, the LLM was unable to detect reliable patterns, even with a context-aware prompt. This points to the importance of refining study protocols, such as incorporating clearer guidance, structured activity periods, or enhanced data monitoring to generate high quality data that can meaningfully inform the recovery process.
\end{itemize}

\section{Conclusions and Future Work}
\label{sec:conclusion}
This paper introduced the idea of clustering multimodal sensor data by each modality to improve understanding of their impact on the recovery of community-dwelling older adult patients after lower-limb fracture. Clustering by each modality breaks down a high-dimensional clustering problem into smaller and more manageable sub-problems that can be combined to cluster the overall dataset. Context-aware prompting and LLMs were used to infer cluster labels. Visualizations of the clinical scores corresponding to the clusters obtained per modality were then performed. Through statistical testing, it was shown that the clusters and their inferred labels are in agreement with each other and that their meanings can help in understanding the recovery trajectories of patients. It was also found that if a deeper understanding of the collected data is not provided to the LLM, the inferred cluster labels may not be useful from an interpretability perspective. This opens up new avenues to extract more informative and clinically relevant features per modality to improve the dataset’s discriminative power.

This preliminary analysis leads to multiple future directions. The most intuitive direction is to combine the outcomes of multiple clusterings from each modality to perform a final clustering on the overall dataset. This will help interpret those clusters to group patients based on different sensing modalities. A comparison with single-view clustering will highlight the benefits of using multiview clustering in terms of explainability of the dataset. The clinical data could also be combined with sensor data to assess whether that improves clustering performance and contributes to explainability. The data have temporal and value-based dimensions; thus, spatio-temporal clustering algorithms \cite{ansari2020spatiotemporal} should be applied to understand the relationship between time and patterns in sensor data during patients' recovery. Lastly, this idea should be tested on other real-world sensor datasets, and its clinical implications must be further investigated.

\section*{Acknowledgment}
The research work is supported through grants funded by the Natural Sciences and Engineering Research Council of Canada.

%
%
%
\bibliographystyle{IEEEtran}
\bibliography{references}

\end{document}